\documentclass[letterpaper, 10 pt, conference]{ieeeconf}
\IEEEoverridecommandlockouts            	
                                                          
\usepackage{graphics} 	
\usepackage{epsfig} 	
\usepackage{mathptmx} 	
\usepackage{times} 		
\usepackage{booktabs}

\usepackage{cite}
\usepackage{amsmath,amssymb,amsfonts}
\usepackage{algorithmic}
\usepackage{graphicx}
\usepackage{textcomp}
\usepackage{xcolor}
\usepackage[hidelinks]{hyperref}
\usepackage{caption}
\usepackage{subcaption}
\usepackage{enumerate}
\usepackage{multirow}
\usepackage{tabularray}

\title{\LARGE \bf
PIPE: Process Informed Parameter Estimation,\\a learning based approach to task generalized system identification
}

\author{Constantin Schempp and Christian Friedrich$^{1}$
\thanks{$^{1}$ All authors are with the Department of Mechanical Engineering and Mechatronics,
Karlsruhe University of Applied Sciences (HKA), Germany,
{\tt\small \{constantin.schempp, christian.friedrich\}@h-ka.de}}%
}

\begin{document}

\maketitle
\thispagestyle{empty}
\pagestyle{empty}


\begin{abstract}
We address the problem of robot guided assembly tasks, by using a learning-based approach to identify contact model parameters for known and novel parts.
First, a Variational Autoencoder is used to extract geometric features of assembly parts. In addition, we propose a new type of network structure, combining this VAE with physical knowledge through a contact model, allowing the parameters of the model to be derived by the geometric features. The measured force from real experiments is used to supervise the predicted forces, thus avoiding the need for ground truth model parameters.
Although trained only on a small set of assembly parts, good contact model estimation for unknown objects were achieved.
Our main contribution is a novel network structure which allows to estimate contact models of assembly tasks depending on the geometry of the part to be joined.
Where current system identification processes have to record new data for a new assembly process, our method only requires the 3D model of the assembly part.
We evaluate our method by estimating contact models for robot-guided assembly tasks of pin connectors as well as electronic plugs and compare the results with real experiments.
\end{abstract}

\section{Introduction}
\begin{figure}[!ht]
\centering
\includegraphics[scale=1.0]{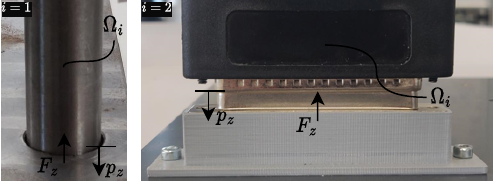}
\caption{Depiction of the variables describing the robot guided assembly task. The first picture shows a dowel pin and the second picture a DSUB37 plug.
$\Omega_i$ is the object at assembly task $i$, $F_z$ the resulting force during assembly and $p_z$ the positon during contact.}
\label{fig:task as a system}
\end{figure}

Robots are increasingly being employed for the automated assembly and disassembly of consumer products such as electronic devices as well as industrial components.
During an assembly process, two or more components come into contact, see \autoref{fig:task as a system}. This results in reaction forces during the joining process.
To ensure a reliable assembly process, force-torque controllers such as admittance \cite{Yoon_Na_Song_2024}, impedance \cite{Song_Kim_Song_2015} or hybrid force-position controllers \cite{Wang_Zou_Su_Luo_Li_Huang_2021} are usually employed.
In order to achieve a reliable assembly process, the controller parameters must be designed depending on the contact dynamics of the respective assembly step.
A model of contact dynamics is necessary to determine control parameters suitably. 
However, a generalization of the model and thus its transferability to similar processes is usually not possible, as only data from a single task is utilized in the design. This implies that the model can only be generalized to unknown input signals, but not to changes in the process information.
\begin{figure}[!ht]
\centering
\includegraphics[scale=1.0]{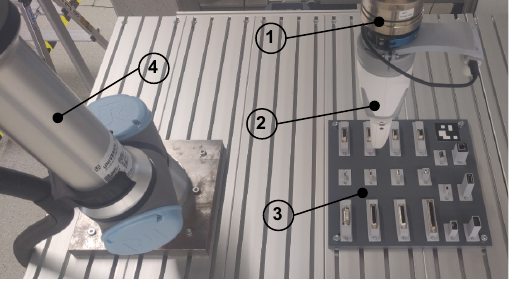}
\caption{Setup of the robot guided assembly task, consisting of (1) force-torque-sensor FTN-AXIA80, (2) gripper SCHUNK EGH 80-IOL-N, (3) task board and (4) UR5e robotic arm.}
\label{fig:aufbau}
\end{figure}
Therefore, if the geometry of the component to be assembled changes, a new model of the contact dynamics has to be determined and the controller has to be adapted accordingly.
This is time-consuming and limits an automated and flexible controller design.
In contrast, we utilize the fact that process information contains data that defines the properties of the assembly task.
Depending on the specific process, various modalities are suitable, such as 3D data (geometry of the object to be joined), color images (environmental features) or material properties such as stiffness and elasticity. 
Embedding this data would make an important contribution to increasing the flexibility of robot systems with regard to force controller design in contact-rich assembly. Therefore, we propose a method which estimates contact models depending on the geometry of the assembly part,
thereby being able to generalize within a given task domain.

\section{Related Work}
Conventional system identification methods capture the behavior of a system by observing its inputs and corresponding outputs.
After selecting a suitable model, the input and output data are used in an optimization procedure to determine the model parameters that best fits the system dynamics \cite{Nelles_2020}.

For solving assembly tasks conventionally, adaptive control is widely deployed \cite{Villa_Mobedi_Ajoudani_2022, Ortner_Gadringer_Gattringer_Mueller_Naderer_2020} to ensure smooth insertion. The controller parameters are set manually for the required task.

Purely neural network based approaches, such as using MLPs \cite{Forgione_Piga_2021}, CNNs \cite{Andersson_Ribeiro_Tiels_Wahlström_Schön_2019} and Autoencoders \cite{Masti_Bemporad_2018} for model identification can be seen as black box approaches where no
prior knowledge is needed. Like in the classical approach, they are optimized on a set of input output data to minimize the error between
real and estimated model response.

Hybrid methods combine conventional modeling schemes and data based approaches to estimate the system dynamics \cite{Bartholdt_Wiese_Schappler_Spindeldreier_Raatz_2021}. For example Physics Informed Neural Networks (PINNs)\cite{PINN} incorporate the residuals of the differential equation, boundary conditions and initial conditions
into the loss function of a neural network to capture the physics. 
Hereby, the inverse problem is described as estimating the unknown parameters and boundary conditions of a physical problem.
Therefore, solving the inverse problem of a PINN can be regarded as a parameter estimation of governing equations.
PINNs have good extrapolation capability and are able to be trained with sparse measurements.
However, since they can only get solutions for a single computational domain, they have to be retrained for new task settings \cite{ScientificMachineLearning}. In \cite{PIPN}, pointnet \cite{PointNet} is used as a PINN to capture geometric features of computational domains, enabling it to solve Navier-Stokes
equations for irregular geometries without retraining. But, after training it is not possible to change the excitation of the differential equation, 
as it is ingrained in training as a loss. This is a general limitation of the PINN approach.

Statistical generalization approaches, like Gaussian Process Regression (GPR)\cite{GPR} or Locally Weighted Regression (LWR) \cite{LWR}, 
can generalize to new assembly tasks when integrated in a task parameterized setting. However, the distribution of the example trajectories and corresponding queries
strongly influence the success of the generalization. Queries have to be equally distributed and recorded data has to transition smoothly between queries \cite{Kramberger_Gams_Nemec_Schou_Chrysostomou_Madsen_Ude_2016}. In addition, the queries are defined manually, which contradicts an automated application.
Task parameterized GPR may be possible for peg in hole tasks, but not for more complex assembly tasks.
All approaches, including classical, neural network-based, hybrid, and task parameterized GPR, exhibit poor contact model generalization when faced with new assembly tasks.
In the following chapter, a new hybrid approach is presented that makes use of the objects geometry to estimate model parameters, allowing generalization
within a given task domain. 

In this paper we present a novel hybrid approach that
\begin{itemize}
\item is able to estimate contact model parameters based on geometric features
\item can generalize within a given task domain
\item allows insight into the model behavior
\item enables integration of prior knowledge
\end{itemize}

\section{Method}
\begin{figure}[!t]
\centering
\includegraphics[scale=1.0]{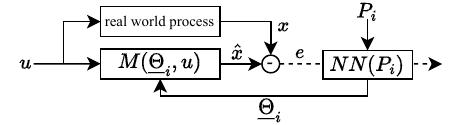}
\caption{Overview of the approach. $P_i$ corresponds to process information of different assembly tasks $i$. $NN(\cdot)$ is the neural network mapping the process information to model
parameters $\underline{\Theta}_i$. $u$ and $x$ describe the system input and state variable respectively.}
\label{fig:method overview}
\end{figure}

The basic idea is to determine the parameters of a model indirectly using the process information $P$, rather than directly with input and output data.
Since $P$ can be high dimensional data, it is not directly transferable into a parametric model representation.
To solve this issue, we use a neural network to estimate model parameters given current process information $NN(P)\rightarrow\underline{\Theta}$.
Where $\dim(\underline{\Theta})$ depends on the model structure $M(\cdot)$ representing the assembly task.
This yields a method which is capable of generalizing and thus identifying contact model parameters depending on process information within the related task domain.
By change of process information, conventional methods require new measurements to fit a model to these new conditions. 
In our approach, the model can generalize within the trained task domain, avoiding renewed data acquisition.
An overview of the general method is given in \autoref{fig:method overview}.

\begin{figure*}
\begin{tabular}{ccccc}
\includegraphics[width=0.1\textwidth]{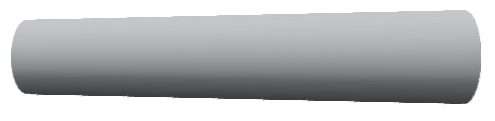} &
\includegraphics[width=0.1\textwidth]{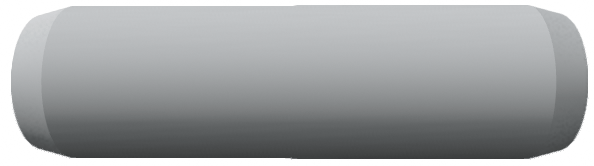} &
\includegraphics[width=0.1\textwidth]{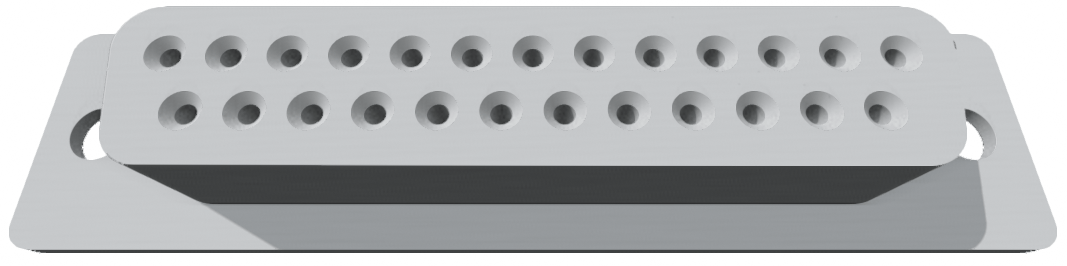} &
\includegraphics[width=0.08\textwidth]{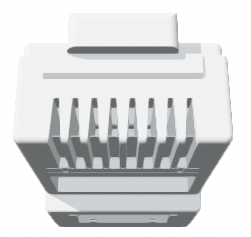} &
\includegraphics[width=0.08\textwidth]{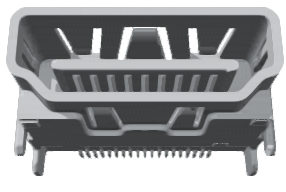} \\ 
\includegraphics[width=0.18\textwidth]{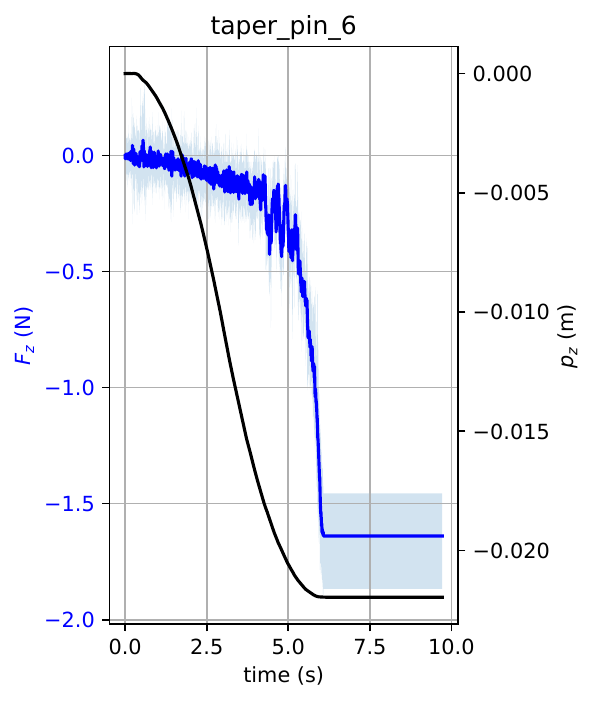} &
\includegraphics[width=0.18\textwidth]{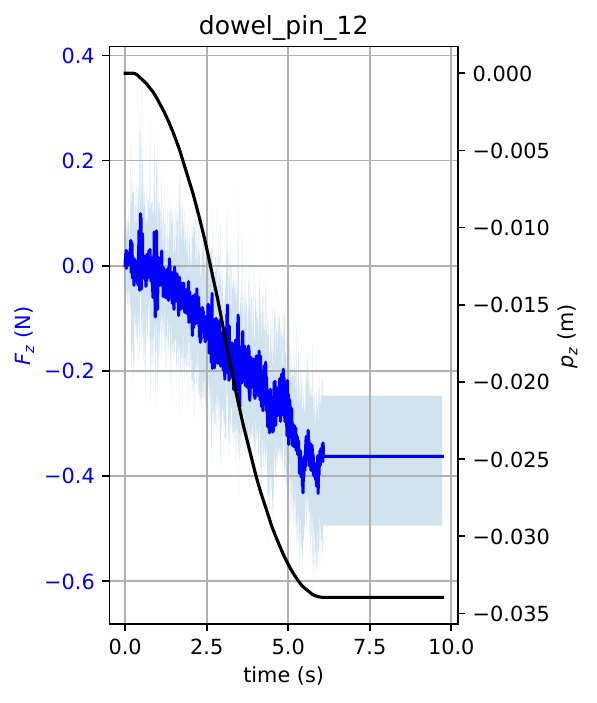} &
\includegraphics[width=0.18\textwidth]{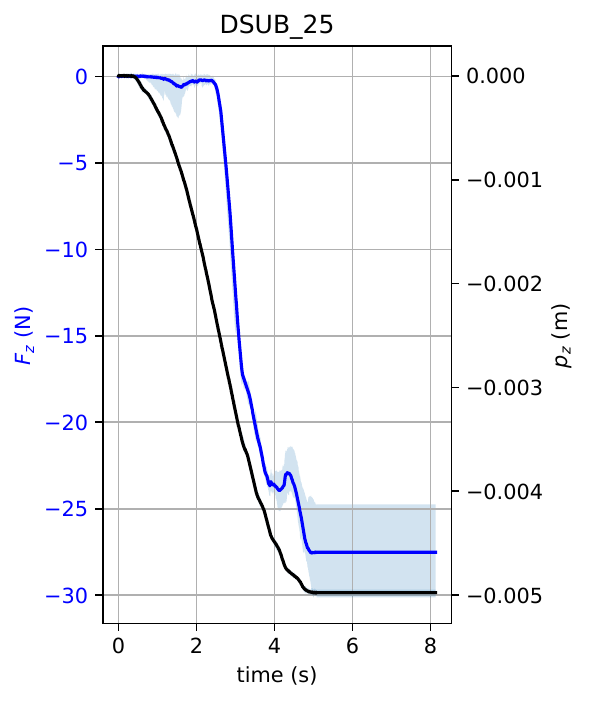} &
\includegraphics[width=0.18\textwidth]{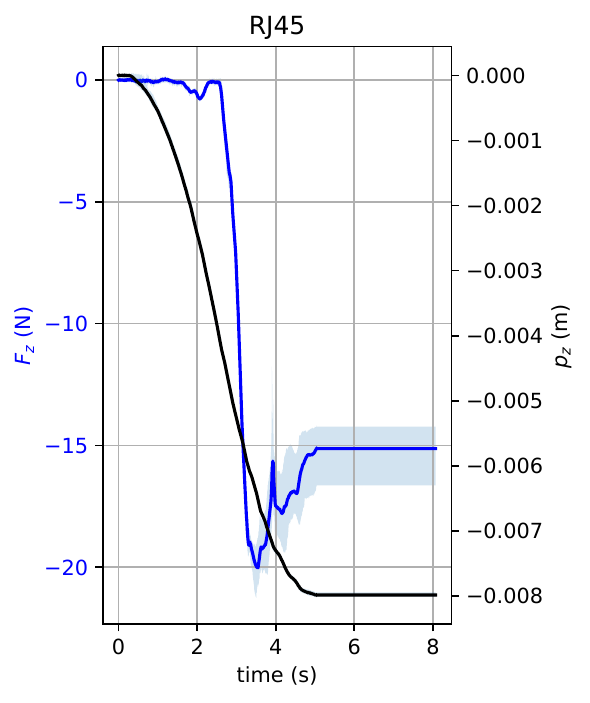} &
\includegraphics[width=0.18\textwidth]{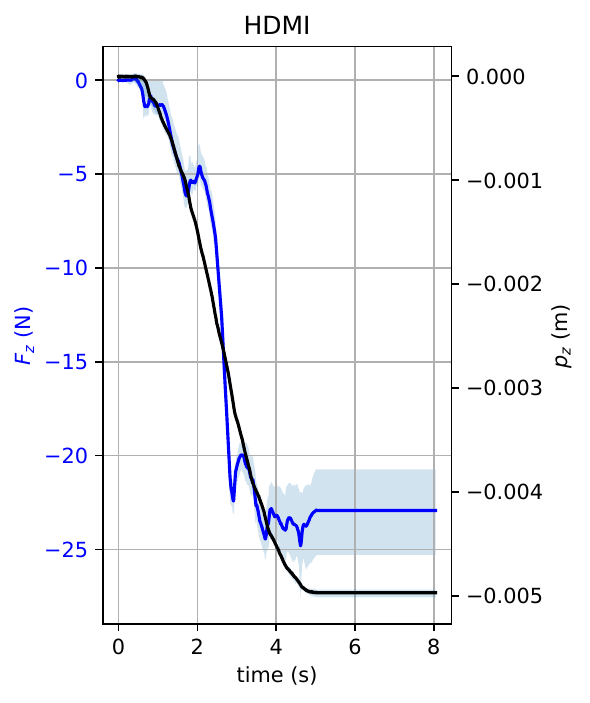} \\
\end{tabular}
\caption{Selection of training data examples used in different task domains. In each column, the mesh of the object and the measured force $F_z$ in blue and position profile $p_z$ in black during assembly is shown. The shaded area depicts the variance and the solid line the arithmetic mean of the $10$ different assembly tasks.}
\label{fig:training data}
\end{figure*}

\subsection{Task as a System}
In order to estimate model parameters and therefore system dynamics, the assembly task has to be represented as a system. Therefore we have to:
\begin{enumerate}[(i)]
\item define the state variable $x$ (the quantity whose dynamic behavior is to be modeled)
\item define the system input $u$ (the quantity exciting the state variable)
\item define the process information $P$ influencing the behavior of the state variable
\item choose a parametric dynamic model $M(\underline{\Theta}, u)$
\end{enumerate}
where $x$ and $u$ can be multivariate when used in a MIMO setting. $P=\{\rho_1, ..., \rho_n\}$ is a collection of process information influencing the state variable and
$M(\underline{\Theta},u)$ can be any appropriate model, here described as a linear differential equation:
\begin{align}
\sum_{i=0}^n a_i \cdot x^{(i)}(t) = \sum_{j=0}^m b_j \cdot u^{(j)}(t)
\end{align}
We summarize all coefficients in a vector
\begin{align}
\underline{\Theta} = [a_0, \ldots, a_n, b_0, \ldots, b_m]
\end{align}

\subsection{General Network Structure}
The network consists of three main components: Encoder $E$, regression head $R$ and Dynamic Model $M$.
First, $P$ gets encoded to relevant features.
Following, the features are the input to the regression head to estimate $\underline{\Theta}$.
Finally, $\underline{\Theta}$ and $u$ are used to compute the system response $\hat{x}$ of $M(\underline{\Theta}, u)$.
\begin{align}
E(P) \rightarrow z; \qquad R(z) \rightarrow \underline{\Theta}; \qquad M(\underline{\Theta}, u) \rightarrow \hat{x}
\end{align}
We summarize encoder and regression head as one network $NN(P) \equiv R(E(P))$.

\section{Application using a robot guided assembly task}
We demonstrate the method on robot guided assembly tasks, carried out by an UR5e from Universal Robots.
The robot performs various assembly tasks, such as plugging dsub-connectors with different number of pins, and insertion of taper and dowel pins of different diameter.
A force-torque sensor (FTN-AXIA80) is attached to measure the forces during assembly.
Our goal is to estimate the forces that occur when the object to be assembled changes.
This involves learning the relationship between object geometry and assembly force.
The setup is shown in \autoref{fig:aufbau}.

\subsection{Task as a System}
\label{subsec:task_as_a_system}
We choose the force $F_z$ in $z$-direction as state variable as we want to model the dynamic behavior of the measured force during assembly.
The system input $u$ which is exciting the state variable, is therefore the position of the robot end-effector $p_z$ in $z$-direction.
We utilize different objects during the assembly tasks. Therefore, we choose the process information to be the geometries of the objects in point cloud representation $P_i=\Omega_i$.
We use a 2nd order differential equation with varying parameters to describe the robot environment interaction with a mass-damper-spring system. That is, the parameters depend on the current position $p_z$ of the robot end-effector.
So our task as a system can be summarized as follows:
\begin{align}
&x = F_z; \qquad u = p_z; \qquad P = \Omega \\
\label{eq:dynamic_model}
&M(\underline{\Theta}, u) \rightarrow b_0 p_z = a_0\ddot{F_z} + a_1\dot{F_z} + a_2 F_z \\
&\underline{\Theta} = [a_0(p_z), a_1(p_z), a_2(p_z), b_0(p_z)]
\end{align}
\autoref{fig:task as a system} shows the variables in the task setting.

\begin{figure*}[!t]
\centering
\includegraphics[scale=0.66]{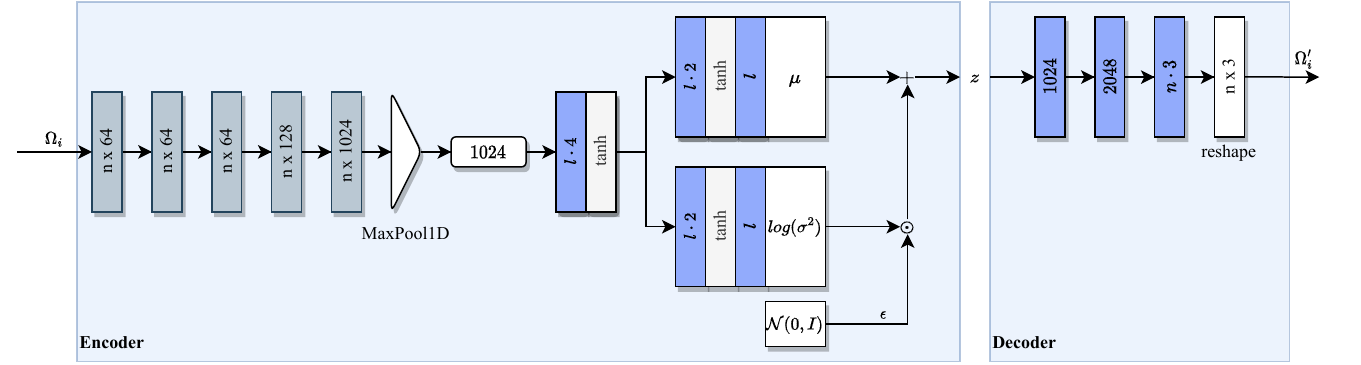}
\caption{Network architecture of the used VAE model. First five blocks indicate 1D convolution with batch normalization and ReLU activation, other blocks indicate fully
connected layer. $\odot$ means element-wise product. $\Omega_i$ is the input geometry and $\Omega_i'$ the reconstructed input.
$l$ stands for the latent-space dimension and $n$ for the number of points in the point cloud.}
\label{fig:vae architecture}
\end{figure*}

\subsection{Training Data \& Preprocessing}
The training data comprises the process information that affects the assembly task, such as the object geometry, and the recorded state variables $x$ and system input $u$, such as the measured forces during assembly and position trajectory.
During assembly we deploy objects from various domains, namely dsub connectors like dsub9 and vga, taper and dowel pins as well as conventional connectors like rj45 and usb, see \autoref{table:training_data}. Each assembly task is repeated $10$ times.
The point clouds are preprocessed by sampling $4096$ points for each CAD object using the sample elimination method \cite{SampleElimination} and aligning the object to the volume center $\underline{0}$.
The position and force profiles are preprocessed to start at zero, i.e. at the beginning, there is no contact between the components.
The final value in the position and force profiles is duplicated and appended, allowing the network to learn steady-state behavior.
Training data is shown in \autoref{fig:training data}.

\begin{figure}[!t]
\centering
\includegraphics[scale=0.7]{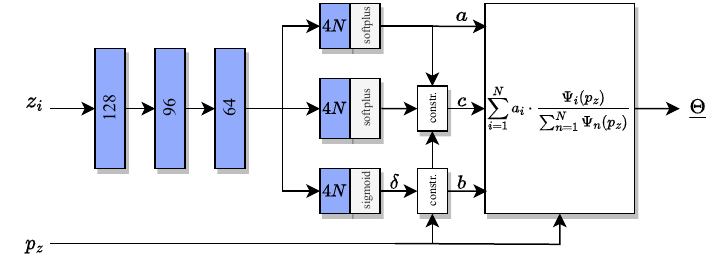}
\caption{Network architecture of the regression head. First three layers are ResNetMLP \cite{he2015deep} blocks, followed by normal MLPs.
$a$, $b$ and $c$ represent respectively amplitude, mean and variance of the gaussian.
$b$ is calculated by constraining the sections $\delta$ on the value range of $p_z$. }
\label{fig:regression head}
\end{figure}

\subsection{Network Architecture}
The Network architecture consists of three components. The first one is a Variational Autoencoder (VAE), which encodes high dimensional process information
to a lower dimensional latent representation. The second component is a regression head, which makes use of the latent space to estimate the parameters of the dynamic model described in \ref{subsec:task_as_a_system}. The dynamic model is then employed to determine the system response when excited with the input $u$.
Network architectures are depicted in \autoref{fig:vae architecture} and \autoref{fig:regression head}.

\subsubsection{VAE}
We use a VAE as pretraining method to extract features from the data describing the process.
VAE was chosen because of its embedded distribution, which enables the model to sample data from the same embedding distribution as the input data, thus making it possible to sample continuous features of unknown assembly parts \cite{kingma2022autoencoding}.
We use a pointnet encoder structure to encode the high dimensional point clouds into a latent space representation.
For the dimension of the latent space we chose $l=32$.

\begin{figure}
\centering
\includegraphics[scale=0.65]{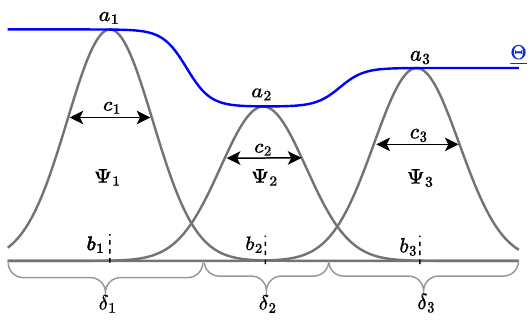}
\caption{
Representation of the estimated parameter profile $\underline{\Theta}$.
Local models $\Psi$ are centered at $b$ with variance $c$ and constrained to lie within the value range of $p_z$.}
\label{fig:local_models}
\end{figure}

\subsubsection{Regression Head}
The regression head uses the latent-space representation $z$ as input to predict the variable parameters $\underline{\Theta}$ of the dynamic model.
This is done by estimating local models at centers $b$ which are spaced over the value range of $p_z$.
These local models are represented by $N$ Gaussian $\Psi_n$ with variance $c$ and amplitude $a$, see \autoref{fig:local_models}.
We use softplus as activation for estimating the amplitude and variance to ensure positive non-zero value.


We observed that directly estimating $b$ leads to badly distributed local models.
Therefore, we first estimate sections $\underline{\delta}$ in which the models are centered.
These are then scaled so that their sum equals the maximum absolute value of $p_z$:
\begin{align}
\underline{\delta_s} = \frac{\underline{\delta}}{\sum_i^N\delta_i} max(|p_z|)
\end{align}
$\underline{\delta}$ is a $N$ dimensional column vector, with $N$ the number of local models and $\underline{\delta_s}$ are the scaled sections.
Multiplication with $Q$ results in centers $b$ of local models constrained to the value range of $p_z$:
\begin{align}
b &= \underline{\delta_s}^T \cdot Q \\
Q_{N,N} &= U_{N,N} - \frac{1}{2}I_{N,N}
\end{align}
$U$ is an upper triangular matrix of dimension $N\times N$ consisting of ones and $I$ is the identity of dimension $N\times N$. Thus, $Q$ is an upper triangular matrix consisting of ones with the main diagonal set to $\frac{1}{2}$.
This encourages a better distribution of local models over $p_z$.

We have to restrict the estimated variances, so that the influence of the local models does not extend over the entire value range of $p_z$.
This is achieved using (\ref{eq:gaussian}) to calculate $c$, so that $\Psi(\underline{\delta_s}^TU)=\xi$ applies.
$\xi$ is the value of the local model at the edges of $\underline{\delta_s}$.
This gives us the possibility to constrain the area of influence of the local models.
We use $\xi=a/20$ for $c_{min}$ and $\xi=a/3$ for $c_{max}$.

Using the estimated amplitudes $a$, centers $b$ and variances $c$, $N=10$ Gaussian are computed, normalized and summed to yield the parameter profile $\underline{\Theta}(p_z)$:
\begin{align}
\underline{\Theta}(p_z) &= \sum_{i=1}^N a_i \cdot \frac{\Psi_i(p_z)}{\sum_{n=1}^N \Psi_n(p_z)} \\
\Psi_i(p_z) &= e^{\frac{-(p_z-b_i)^2}{2c_i^2}} \label{eq:gaussian}
\end{align}
The normalization ensures that the value of $\underline{\Theta}(p_z)$ always runs within the limits given by the value of the 
adjacent support values and at the same time a monotonicity of the support values also causes a monotonic course of $\underline{\Theta}(p_z)$.

\begin{figure}[!t]
\centering
\includegraphics[width=0.5\textwidth]{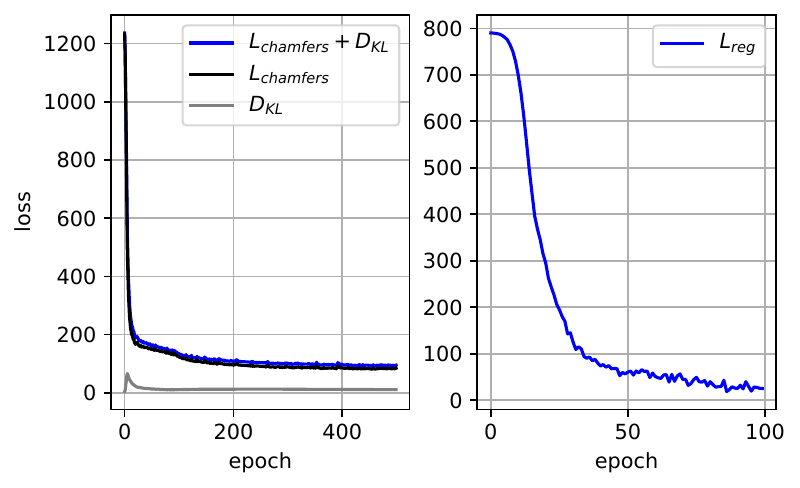}
\caption{Losses during training of the VAE (left) and regression head (right). $D_{KL}$ represents the KL divergence.}
\label{fig:losses}
\end{figure}

\begin{figure*}[!t]
\centering
\resizebox{\textwidth}{!}{%
\begin{tblr}{rowspec={Q[c,m] Q[c,m] Q[c,m] Q[c,m] Q[c,m] Q[c,m]}, colspec={Q[1.8cm] Q[2.85cm] Q[2.85cm] | Q[1.8cm] Q[2.85cm] Q[2.85cm]}}
\SetCell[c=3]{} taper8, $\text{rms}=0.57$ N, $FIT(F)=43.31$\%, $FIT(\dot{F})=50.76$\% & & & \SetCell[c=3]{} dsub26, $\text{rms}=11.92$ N, $FIT(F)=42.68$\%, $FIT(\dot{F})=65$\% & & \\ \hline

{\includegraphics[width=0.08\textwidth]{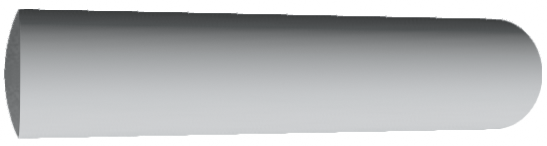}\\($a_1$)} & 
\SetCell[r=3]{}{\includegraphics[width=0.15\textwidth]{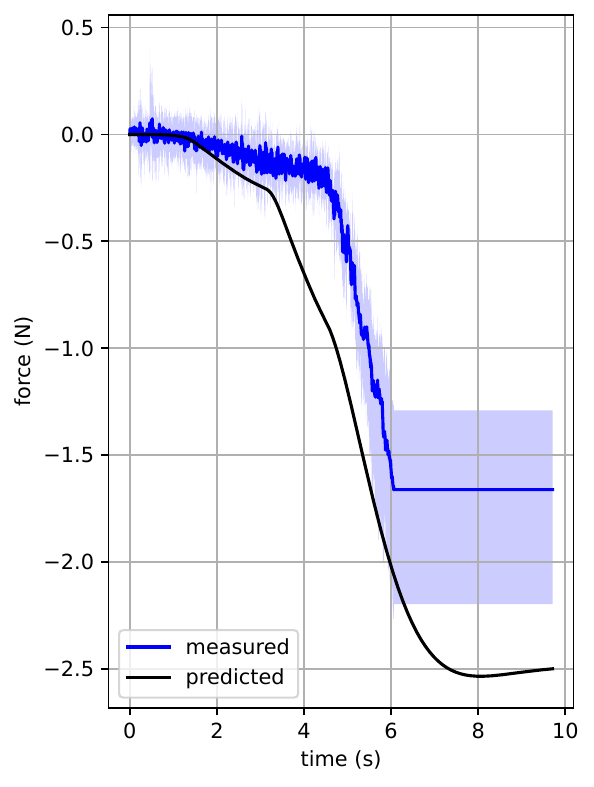}\\($d_1$)} &
\SetCell[r=3]{}{\includegraphics[width=0.15\textwidth]{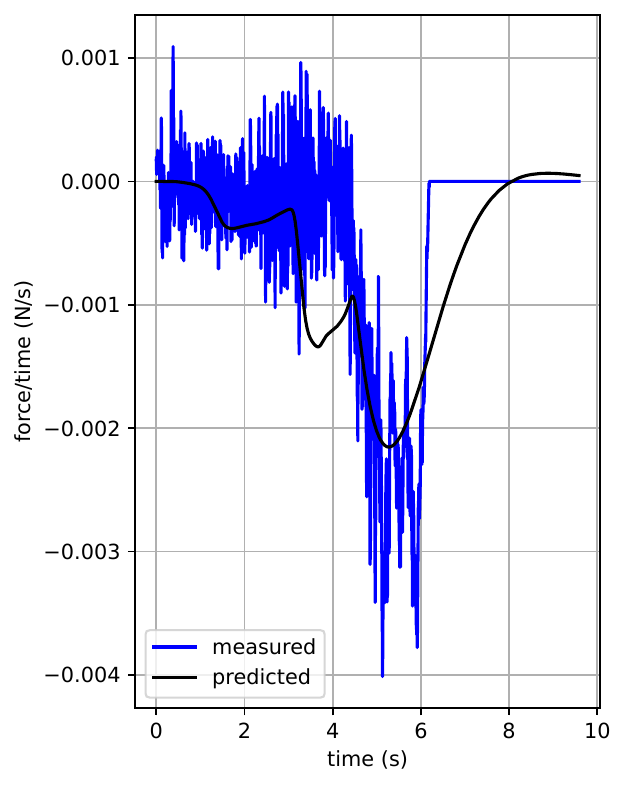}\\($d_1$)} &
 
{\includegraphics[width=0.1\textwidth]{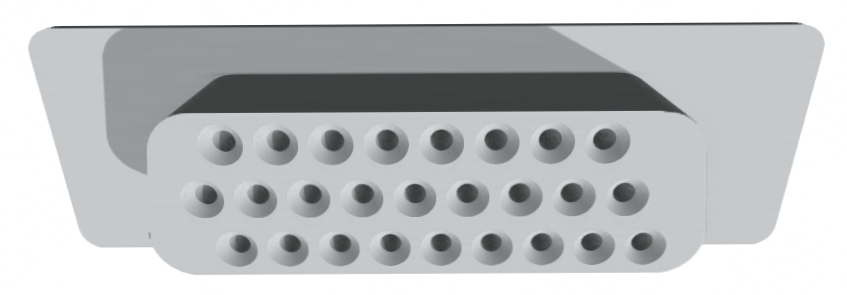}\\($a_2$)} & 
\SetCell[r=3]{}{\includegraphics[width=0.15\textwidth]{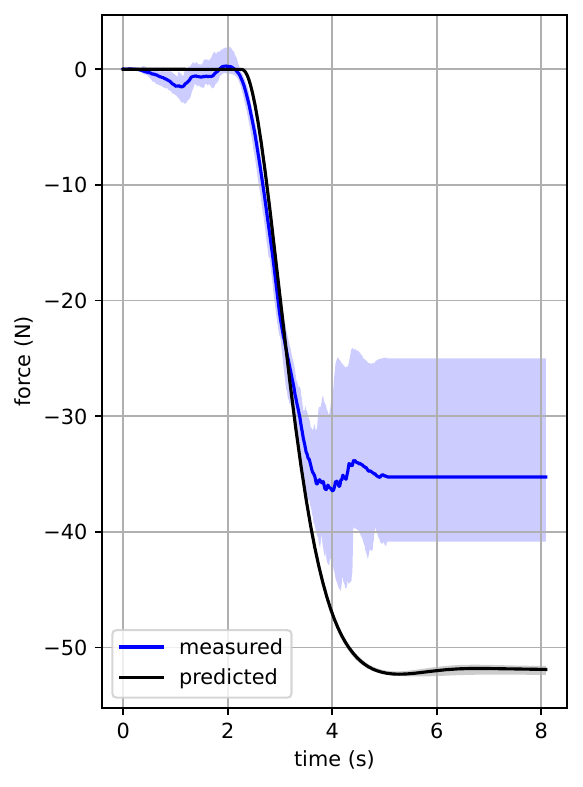}\\($d_2$)} & 
\SetCell[r=3]{}{\includegraphics[width=0.15\textwidth]{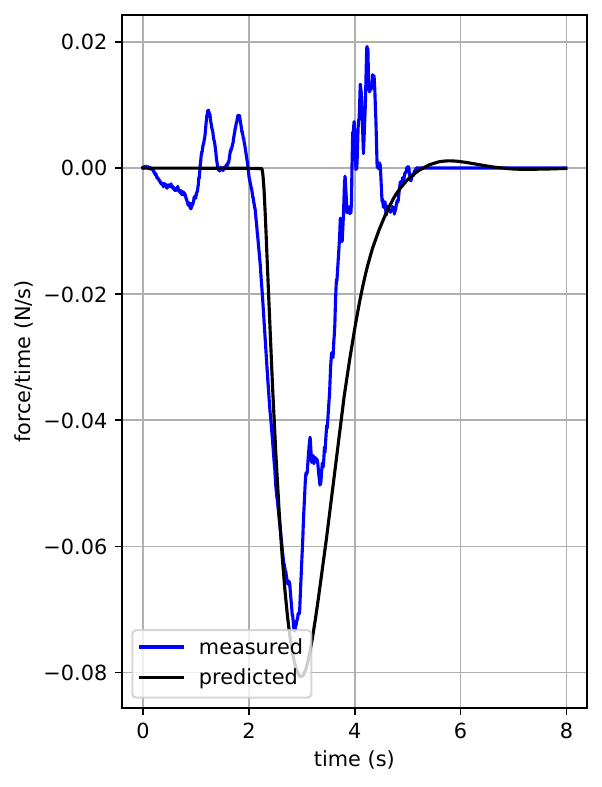}\\($d_2$)} & \\

{\includegraphics[width=0.1\textwidth]{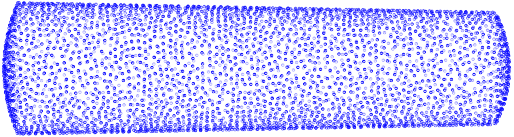}\\($b_1$)} &
& &
{\includegraphics[width=0.1\textwidth]{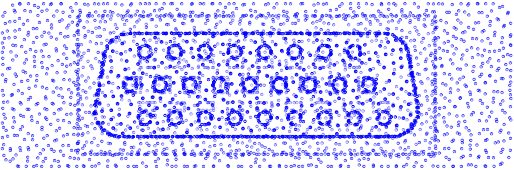}\\($b_2$)}& & &\\
{\includegraphics[width=0.1\textwidth]{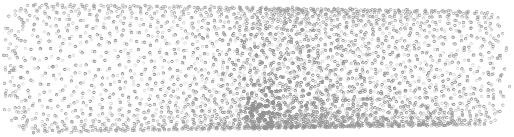}\\($c_1$)} &
& &
{\includegraphics[width=0.1\textwidth]{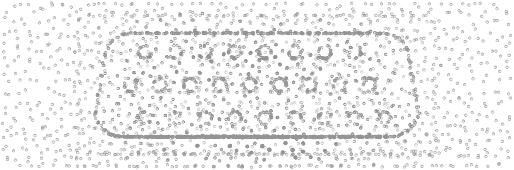}\\($c_2$)}
\end{tblr}}
\caption{Results of inference with unseen objects. ($a$) depicts the mesh of the object, ($b$) the input point cloud to the VAE, ($c$) the reconstructed point cloud using the encoded
latent space and ($d$) the measured force during assembly in blue and the predicted force by exciting the estimated dynamic model in black.}
\label{fig:evaluation_test_data}
\end{figure*}

\subsubsection{Dynamic Model} \label{subsec:dynamic_model}
For the dynamic model we choose a 2nd order ODE with varying parameters $\underline{\Theta}$, which is excited by $p_z$, see (\ref{eq:dynamic_model}).
The choice of model depends on the complexity of the state variable.
If prior knowledge of the system is available, then a more appropriate model can be selected.
Because we are using a superposition of local models, we can adapt the model complexity by increasing the number of local models $N$.
Solving this equation yields the estimated force $\hat{F}$ during assembly. 

\subsection{Training} \label{subsec:training}
Training consist of a two-stage approach. First, the VAE is trained to reconstruct the input geometry $\Omega$.
The VAE was trained for 500 epochs with a batchsize of $10$ and learning rate of $10^{-4}$ using the Adam optimizer.
The Chamfer distance is used as reconstruction loss, which is computed by summing the squared
distances between nearest neighbor correspondences of two point clouds
\begin{align}
L_{chamfer} = \sum_{x \in X} \underset{y \in Y}{\text{min}} \Vert x-y \Vert_2^2 + \sum_{y \in Y} \underset{x \in X}{\text{min}} \Vert x-y \Vert_2^2 ,
\end{align}
where $X$ and $Y$ are the two point clouds, while $x$ and $y$ represent each point in their respective point cloud. 
Afterwards, we freeze the encoder and append the regression head and the dynamic model.
The regression head is then trained for another $100$ epochs with batchsize $10$ and learning rate $2 \cdot 10^{-4}$.
As a loss function we use the MSE between measured $F$ and estimated force $\hat{F}$ during assembly
\begin{align*}
L_{reg} = \frac{1}{n} e^Te
\end{align*}
where $e=(F - \hat{F})$ and $e \in \mathbb{R}^{n}$. Here too we apply Adam for optimization.
For each task domain, we use the training data listed in \autoref{table:training_data}. 
\begin{table}
\renewcommand{\arraystretch}{1.3}
\caption{training data used for each task domain. The number behind dsub depicts the number of contacts, while the number behind pins depicts the nominal diameter in mm.}
\label{table:training_data}
\centering
\begin{tabular}{@{}llll@{}}
\toprule
domain                                                            & train                                                                                                                                & val                                                                 & test                                                                  \\ \midrule
dsub                                                              & \begin{tabular}[c]{@{}l@{}}dsub9, dsub15, dsub25,\\ dsub37, dsub44\end{tabular}                                                      & vga                                                                 & dsub26                                                                \\
pins                                                              & \begin{tabular}[c]{@{}l@{}}taper6, taper10, taper14,\\ dowel5, dowel12\end{tabular}                                                  & \begin{tabular}[c]{@{}l@{}}taper12,\\ dowel8\end{tabular}           & \begin{tabular}[c]{@{}l@{}}taper8,\\ dowel6\end{tabular}              \\
\begin{tabular}[c]{@{}l@{}}conventional\\ connectors\end{tabular} & \begin{tabular}[c]{@{}l@{}}dsub9, dsub15, dsub25,\\ dsub37, dsub44, dvi,\\ hdmimini, hdmi, rj45,\\ usba, usbc, usbmicro\end{tabular} & \begin{tabular}[c]{@{}l@{}}vga,\\ hdmimicro,\\ usbmini\end{tabular} & \begin{tabular}[c]{@{}l@{}}displayport,\\ dsub26,\\ usbb\end{tabular} \\ \bottomrule
\end{tabular}
\end{table}
The course of the losses is shown in \autoref{fig:losses}.

\subsection{Evaluation}
We evaluate our method by training a model for each domain and inferring with the unseen objects from the test data mentioned in \autoref{table:training_data}.
\autoref{fig:evaluation_test_data} shows the results of inference with the unknown plug and taper pin.
The reconstructed point cloud has a mean distance error of $0.012$ mm, indicating that the encoded latent space is a good representation of the input point cloud.
Using the latent space, the parameters of the contact model were estimated, which is capable of representing the behavior of the force during assembly of the object.
For evaluating the estimated contact behavior, the fit ratio $FIT(x)=(1-\text{nmse(x)})\cdot 100$ is calculated, where nmse is the normalized mean squared error.
The general contact behavior during assembly can be inferred from the objects geometry.
This is indicated by a $FIT(\dot{F})$ of $50.76$\% for taper8 (domain pins) and $65$\% for dsub26 (domain dsub) respectively. 
However, an error in form of a scaling factor is still existent, visible in the last section of the estimated force profiles, resulting in $FIT(F)$ of $40.31$\% and $42.68$\%. A possible reason is the lack of training data, which limits the networks generalization capability.
This can be recognized by the fact that the model trained in the conventional connectors domain results in $FIT(F)=56.57$\% and $FIT(\dot{F})=68.69$\% for dsub26.

The rms and relative rms of the predictions are listed in \autoref{table:results}, where the relative rms is calculated by
equal rescaling of the measured forces:
\begin{align}
rms_{rel} = \sqrt{\frac{1}{n}\sum_{i=1}^n \left( \frac{F_i-\hat{F}_i}{\max(|F_i|)} \right) ^2}
\end{align}
This allows a better comparison, since the value range of the different task domains differs.
\begin{table}
\renewcommand{\arraystretch}{1.3}
\caption{results of inference with test data of each task domain.}
\label{table:results}
\centering
\begin{tabular}{@{}lccccc@{}}
\toprule
domain                                                            & \multicolumn{1}{l}{\# train} & \multicolumn{1}{l}{\# val} & \multicolumn{1}{l}{\# test} & \multicolumn{1}{l}{$rms$ [N]} & \multicolumn{1}{l}{$rms_{rel}$ [N]} \\ \midrule
dsub                                                              & 5                            & 1                          & 1                           & 11.92                   & 0.32                         \\
pins                                                              & 5                            & 2                          & 2                           & 0.71                    & 0.66                         \\
\begin{tabular}[c]{@{}l@{}}conventional\\ connectors\end{tabular} & 12                           & 3                          & 3                           & 11.16                   & 0.57                         \\ \bottomrule
\end{tabular}
\end{table}
\section{Conclusion \& Future Work}
In this work, we have demonstrated the feasibility of estimating contact model parameters for robot guided assembly tasks based on the geometry of the components to be assembled.
This was achieved by deriving the model from learned contact behavior of assembly tasks using a novel network structure.
Our approach integrates a conventional dynamic model, which provides insights into the learned contact behavior. This is not possible with black box approaches.
Additionally, the integration of prior knowledge is possible by appropriately setting the centers of the local Gaussian or by choosing an overall
suitable dynamic model $M(\cdot)$.
By embedding a contact model in the network structure, and deriving its parameters based on the geometry of the object, our method combines the dynamic informed property of PINNs with the generalization ability of statistical approaches.
Although our method delivers good results, it also has its limitations. Since currently only the geometry of the component to be assembled is considered, imprecise models are estimated for components of the same geometry consisting of different materials.
In order to address these limitations, future work will explore the combination of additional process information in latent space to account for material properties in addition to geometry.
Furthermore, we will use the estimated dynamic model of the assembly task to perform an automatic controller design 
and transfer the method to other use-cases in order to test the applicability.

\addtolength{\textheight}{0cm} 


\bibliography{IEEEabrv, mybibfile.bib}

\end{document}